\documentclass[journal,twocolumn]{IEEEtran}
\usepackage[utf8]{inputenc}
\usepackage[sorting=none]{biblatex}
\usepackage{circuitikz}
\usepackage{url}
\usepackage[T1]{fontenc}
\usepackage{amsmath}
\usepackage{todonotes}
\usepackage{hyperref}
\usepackage{graphicx}
\usepackage{lipsum}
\usepackage{pict2e}
\usepackage{geometry}
\usepackage{fixltx2e}
\title{A Model Based Approach to Synthetic Data Set Generation for Patient-Ventilator Waveforms for Machine Learning and Educational Use}

\author{
    \IEEEauthorblockN{
        A. van Diepen\IEEEauthorrefmark{1},
        T. H. G. F. Bakkes,
        A. J. R. De Bie,
        S. Turco,
        R. A. Bouwman,
        P. H. Woerlee,
        M. Mischi
    }

        \IEEEauthorrefmark{1} Corresponding author: \emph{a.v.diepen@tue.nl, Dept. of Electrical Engineering, Eindhoven University of Technology, 5612 AZ, Eindhoven, the Netherlands}
    }
    
\begin{document}
\maketitle
\begin{abstract}
 Although mechanical ventilation is a lifesaving intervention in the ICU, it has harmful side-effects, such as barotrauma and volutrauma. These harms can occur due to asynchronies. Asynchronies are defined as a mismatch between the ventilator timing and patient respiratory effort. Automatic detection of these asynchronies, and subsequent feedback, would improve lung ventilation and reduce the probability of lung damage. Neural networks to detect asynchronies provide a promising new approach but require large annotated data sets, which are difficult to obtain and require complex monitoring of inspiratory effort. In this work, we propose a model-based approach to generate a synthetic data set for machine learning and educational use by extending an existing lung model with a first-order ventilator model. The physiological nature of the derived lung model allows adaptation to various disease archetypes, resulting in a diverse data set. We generated a synthetic data set using 9 different patient archetypes, which are derived from measurements in the literature. The model and synthetic data quality has been verified by comparison with clinical data, review by a clinical expert, and an artificial intelligence model that was trained on experimental data. The evaluation showed it was possible to generate patient-ventilator waveforms including asynchronies that have the most important features of experimental patient-ventilator waveforms.

\end{abstract}

\begin{IEEEkeywords}
patient-ventilator interactions, asynchronies, mechanical ventilation, model based methods, machine learning
\end{IEEEkeywords}

\section{Introduction}
Mechanical ventilation is the most common mean of life support applied at the intensive care unit (ICU) \cite{wunsch2013icu}. Mechanical ventilation supports ventilation in many patients after major surgery or in critically ill patients with respiratory failure such as during the recent COVID-19 pandemic \cite{slutsky_ventilator-induced_2013, gattinoni2020covid}.  The goal of mechanical ventilation is to maintain gas exchange that sustains life while minimizing ventilator-induced lung injury (VILI) and work of breath.  Different modes can be applied to mechanically ventilated patients. A mandatory mode is often chosen in patients with more severe respiratory problems as it allows clinicians to completely control ventilation in these patients, who therefore require sedatives. A supportive mode of ventilation, such as pressure support ventilation (PSV), is preferred when a patient’s pulmonary condition improves. In general, since this mode of ventilation is triggered (initiation of a breath) and cycled (termination of a breath) by the patient, it is usually perceived as more comfortable and allows weaning from sedation and eventually from mechanical ventilation. 

\begin{figure}
    \centering
    \includegraphics[trim={0 1cm 0 1cm},clip,width=\columnwidth]{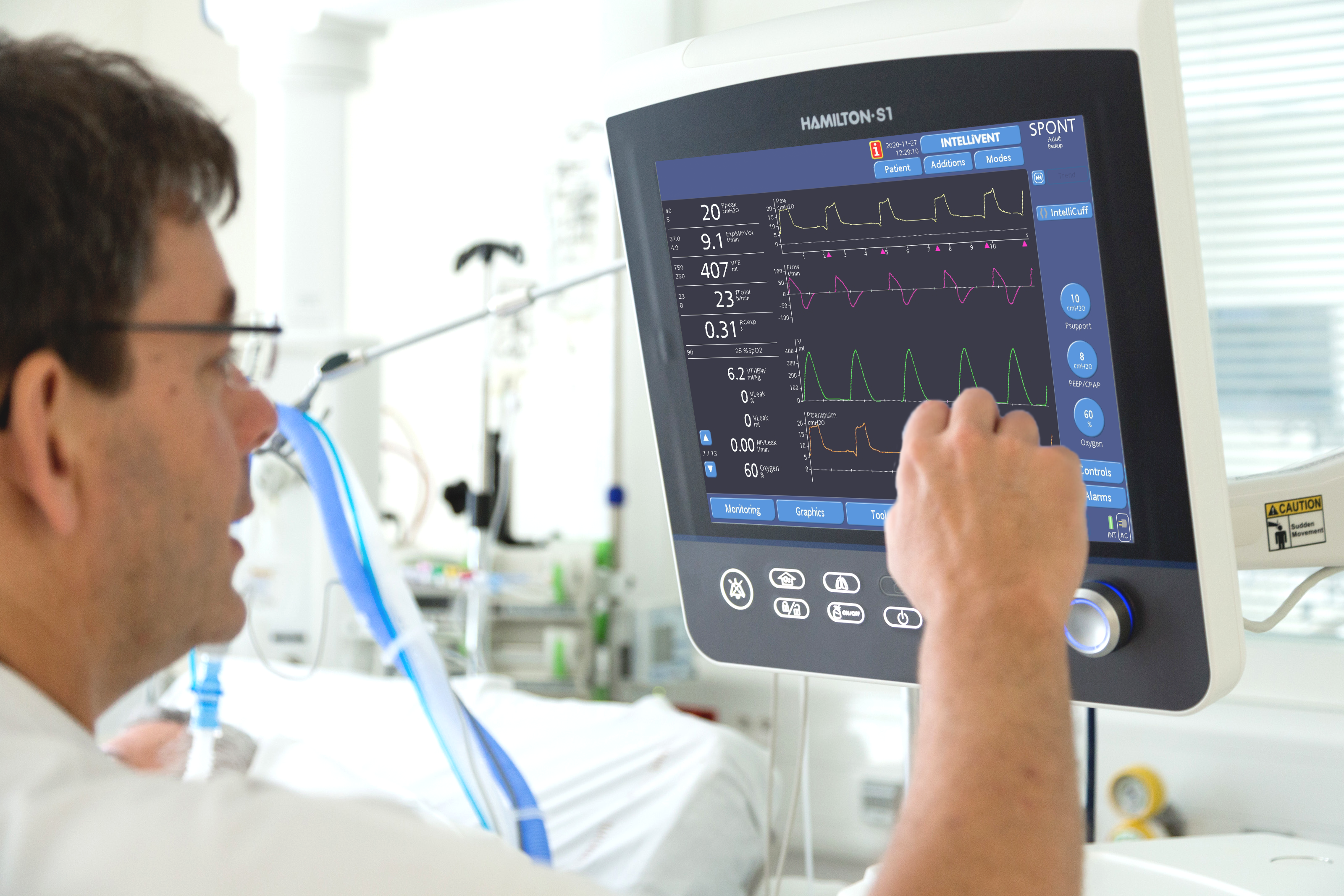}
    \caption{Display of a 
    mechanical ventilator on the bedside with volume, flow and pressure waveforms.}
    \label{fig:mechventwav}
\end{figure}

Mismatches between the patient and the mechanical ventilator during both modes of ventilation can adversely affect the objective of minimizing VILI and work of breath. These so-called patient-ventilator asynchronies are associated with worse outcomes such as discomfort and increased mortality.  However, a direct causal relationship has not yet been scientifically established, nor have we been able to demonstrate if a reduction of asynchronies results in better outcomes \cite{de2019patient, blanch2015asynchronies}. The identification and quantification of asynchronies are therefore crucial to clarify if asynchronies are a direct causative factor. However, detecting patient-ventilator asynchronies by trained clinicians is extremely challenging. The recognition of asynchronies based on the bed-side analysis of waveforms (Figure \ref{fig:mechventwav}) is difficult. Also, trained clinicians are not able to continuously assess these waveforms while asynchronies can occur at any moment. Continuous monitoring with computer algorithms can overcome these barriers and help us to detect, analyze, and maybe even predict asynchronies.

Studies indicate that machine learning algorithms can be used to autonomously detect asynchronies \cite{de2019patient}.
However, classic machine learning approaches are presently insufficiently accurate for practical application \cite{zhang2020detection}.
Neural networks provide an interesting opportunity to elevate the quality of automated asynchrony detection.

Unfortunately, neural networks are notorious for requiring large amounts of labeled data for training, of which there is a shortage~\cite{gholami2018replicating}.
Data acquisition in the field is expensive and greatly complicated by the need for advanced monitoring and by regulatory issues.
Apart from the acquisition difficulties, labeling has to be performed manually by an expert, which is labor-intensive, and prone to errors and ambiguities \cite{zhang2020detection}.
The scarce labeled data that are available do not contain sufficient examples for training and independent testing.
Further complications arise from the variety of asynchronies, which manifest themselves differently in the data.
On top of this, various disease archetypes also alter the measurement data and the distribution of asynchronies throughout the data, further stretching the need for a large and diverse data set.

A common approach when dealing with insufficient data is to augment measurements with synthetic simulations.
In computer vision, researchers have successfully applied this technique on multiple occasions~\cite{gaidon2016virtual,eggert2015benefit, abt2014plea,ghorbani2019dermgan}. 
The advantages of synthetic data are that they are generated by an independent source, labeling is always accurate, and there is full control over the data generation process.
However, to obtain accurate and sufficient synthetic data, precise yet fast simulation models are imperative.
The goal of this study was to investigate the feasibility of generating a labeled synthetic data set of patient-ventilator waveforms for training and testing machine learning algorithms and use in education.

Departing from a validated nonlinear one-compartment lung model by Athanasiades et al. ~\cite{athanasiades2000energy}, this work derives a fast executable lung model with limited loss of accuracy and extends this lung model with a simple ventilator circuit.
The parameters used in the model are based on clinical measurements derived in previous studies of mechanically ventilated patients.

We propose to use this model to generate a sufficiently large, rich synthetic data set that can be used for training and testing machine learning, and that has possible applications in education.
The physiological origin of the model allows adaptation to various asynchrony types combined with various disease archetypes, such as chronic obstructive pulmonary disease (COPD), acute respiratory distress syndrome (ARDS), idiopathic fibrosis, and morbid obesity.

The contributions of this work can be summarized as follows:
\begin{itemize}
    \item Adaptation of the nonlinear lung model from Athanasiades et al.~~\cite{athanasiades2000energy} for fast evaluation in an electronic circuit simulator.
    \item Combination of the adapted lung model with a simple mechanical ventilator model to simulate patient-ventilator interactions. This enables automatic timing labeling of the synthetic data and allows modeling different types of asynchronies.
    \item Modelling of different patient archetypes based on experimental data, including various types of lung conditions.
    \item Validation of the proposed model including expert evaluation by visually comparing the synthetic data to clinical data experienced clinicians and validation with a validated machine learning model. 
\end{itemize}

The remainder of this work is structured as follows:
Section~\ref{sec:relatedwork} contains a study of related work and the current state of the art. 
Adaptation of the lung model, as well as the derivation of the ventilator model, and its implementation are described in Section~\ref{sec:model}.
The model validation and the results are presented in Section~\ref{sec:results}.
Section~\ref{sec:discussion} provides an in-depth discussion, detailing the implications and limitations of our approach as well as expert opinions.
Finally Section~\ref{sec:conclusion} concludes this work.

\section{Related Work} \label{sec:relatedwork}
An approach based on deep learning and neural networks is promising for detecting asynchronies \cite{loo2018machine, subira2018minimizing}.
Instead of a designer deriving rules, like in rule-based algorithms, neural networks derive their own classification mechanisms.
However, neural networks are known for requiring many labeled training examples to reach good performance, while preventing overfitting. 

To the best of our knowledge, the usage of virtual models to generate patient-ventilator waveforms for asynchrony detection and machine learning training has not been studied.
Lino et al. \cite{lino2016critical} review the currently available virtual mechanical ventilator simulators that are focused on educational purposes.
Holanda et al. \cite{holanda2018patient} simulate mechanical ventilator waveforms with asynchronies solely for educational purposes.
The advantage of simulating a lung with software compared to an approach based on mechanical test lungs is that software allows for easy changes of the properties of the lung with no constraints due to ventilator needs.
However, with test lungs it is easier to test the setup for different types of ventilators \cite{richard2002bench}.

Isolated lung models are a well-studied topic in the literature.
Many lung models are based on the one-compartment model, which models the airways and lung as a pipe and a balloon \cite{bates_2009}.
We have chosen to base our work on the validated nonlinear one-compartment model proposed by Athanasiades et al. \cite{athanasiades2000energy}, which in its turn is a continuation of the work of Liu et al. \cite{liu1998airway}.
This model is a more accurate, yet also more complex, version of the regular linear one-compartment model.
In contrast to simpler models, it can emulate the collapse of the middle airways, and the turbulence in the upper airways, the small airways, and the lung and chest wall compliance with nonlinear equations, which is important for accurate ventilation modeling.

Although it has been suggested by Athanasiades et al. \cite{athanasiades2000energy} that their work can be used to model different types of respiratory diseases, there has not been any research on this topic. 
Fortunately, there is a vast amount of literature on changes in lung mechanics under the influence of different disease types \cite{parameswaran2006altered, pelosi1997respiratory, zerah1993effects,kallet2003respiratory,farre1998respiratory, guerin1993lung, yu2015simulation, arnal2018parameters}. 
We use this literature to adapt the model by Athanasiades et al. to different disease archetypes since the model parameters represent real physiological properties of the lung.

To model asynchronies, a ventilator model is required that is compatible with the lung model. 
The ventilator model only incorporates the most important features of the ventilator and is kept as simple as possible.
Most properties of the ventilator are well-studied, e.g., by Kaczka et al. \cite{kaczka2001inspiratory}, but may vary between different ventilator brands.
Literature is sufficient to serve as a basis for deriving a simple ventilator model that can be paired with the selected lung model.

\section{Methods}
\label{sec:model}
The first step towards generating synthetic patient-ventilator data is specifying the underlying model. The model is divided into two parts: the lung model and the ventilator model. 
As stated before, the lung circuit is based on the model by Athanasiades et al. \cite{athanasiades2000energy}, and is altered to accommodate fast evaluation. The first part of this section describes the lung model and the alterations.
After this, different sets of parameters for modeling the different patient archetypes are proposed and explained.
The next part introduces the simple ventilator extension and the last part describes how we combine and implement the lung model and the ventilator model to generate the synthetic data including asynchronies and the validation methods.

\subsection{Mechanical-electrical analogies}

\begin{figure}
\begin{center}
\resizebox{0.9\columnwidth}{!}{%
\begin{circuitikz}[american voltages,scale = 0.75, transform shape]
\ctikzset{font=\fontfamily{phv}\selectfont}
\draw
  (6,4) to [short, l = airway opening] (6,4)
  (6,4) to [vR, l_=R\textsubscript{u}] (6,2)
  (6,2) to [vR, l_=R\textsubscript{c}] (6,0) 
  (6,0) to [vR, l_=R\textsubscript{s}] (6,-2) 
  (6,-2) to [short, l = lung pressure](6,-3)
  (6,-4) to [vC, l = C\textsubscript{l}, mirror] (6,-2)
  (6,-4) to [short] (9,-4)
  (9,-2) to [short, l = intrapleural pressure] (9,-4)
  (9,0) to [vC, l = C\textsubscript{c},invert](9,-2)
  (6,0) to [short] (9,0)
  (6,-6) to [vC, l = C\textsubscript{cw}, mirror] (6,-4)
  (2,-6) to [american voltage source, l = PipPEEP] (2, -8)
  (2,-4) to [R, l = R\textsubscript{d}] (2,-6)
  (2,-4) to [short] (6, -4)
  (6,-6) to [short] (7,-6)
  (6,-6) to [short] (5,-6)
  (5,-6) to [C, l = C\textsubscript{ve}] (5,-8)
  (7,-6) to [R, l = R\textsubscript{ve}] (7,-8)
  (7,-8) to [short] (5,-8)
  (6,-10) to [sinusoidal voltage source, l = P\textsubscript{mus}] (6, -9)
  (6,-8) to [short] (6,-9)
  (6,-10) to (6,-10) node[ground]{}
  (2,-8) to (2,-10) node[ground]{};
\end{circuitikz}
}
\end{center}
\caption{The patient lung model which is an adapted version of \cite{athanasiades2000energy}, note that the components have nonlinear dynamics and are coupled with each other through these dynamics. The Kelvin body and the chest wall compliance are switched as compared to Athanasiades et al. \cite{athanasiades2000energy} and an extra resistor $R_d$ and voltage source $PipPEEP$ are added to ensure the correct functional residual capacity.\label{fig:lungmodel}}
\end{figure}
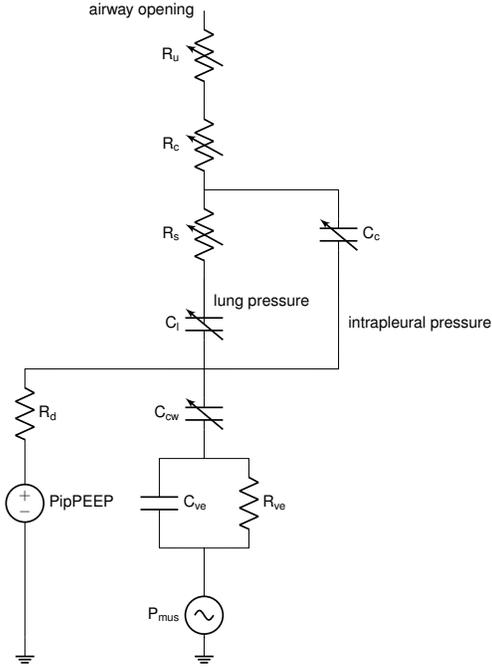


The work of Athanasiades et al. includes a nonlinear one-compartment lumped element model of the lung.
Lumped element models simplify the description of spatial systems by discrete components that can together approximate the behavior of the system. 
In our case, we map a mechanical system (the lung) to the electrical domain (the model).
The translations between mechanical properties and electrical elements are as follows:
\begin{itemize}
    \item The compliance (sometimes called flexibility, a measure of how much a material deforms when stress is applied) is modeled by a nonlinear capacitance in the electrical domain.
    \item The pressure at a certain point in the respiratory system is analog to the voltage at a node in the circuit. A voltage source can thus be seen as a pressure source. Note that usually, when working with both pressure and voltages, we work with voltage and pressure differences. We treat ground as electrical analog for atmospheric pressure.
    \item The resistance to airflow is modeled by a resistor. In general, the smaller the diameter of the tube (or in this case airway) the higher the electrical resistance.
    \item Volume is analog to charge in the electric domain. An important point to keep in mind is that the charge on a capacitor is thus air volume in a specific area of the lung.
    \item The airflow rate is analog to electric current, since this is movement of charge (and thus air volume) from one part of the respiratory system to another.
\end{itemize}

The advantage of this approach is that the rules that apply to the electric domain also translate correctly to the mechanical domain. Both the lung model and the ventilator model make use of this principle.
However, it needs to be noted that not everything is the same and caution must be taken sometimes: for example, negative volume in a physical sense does not exist, while negative charge does. We have taken this into account when using the model.

\subsection{Patient lung model}
By combining the basic elements, a model of the lung is created and shown in Figure \ref{fig:lungmodel}.
The lung model can be divided in four parts:

\begin{itemize}
    \item The top part consists of three resistors to model the airways: the upper airway resistance $R_u$, the collapsible airway resistance $R_c$ and the smaller airway resistance $R_s$. The upper airways are the airways that have cartilage supporting them, due to the large diameter of these airways there is turbulence present at high flows. The turbulence through the large airways is included in the model by adding a Rohrer resistance as described by Athanasiades et al. \cite{athanasiades2000energy}. The small airways are usually supported by the elastic recoil of the lungs, when the lung volume increases the small airways are stretched open, thus decreasing the resistance of the small airways. This effect is embedded in the formula for the small airway resistance. The collapsible airway refers to the middle airway segment, this part of the airway usually has less support and may therefore collapse during expiration. This is an important factor in various disease types, and is usually not included in simpler lung models. A capacitor $C_c$ is added to model the compliance of the collapsible airways.
    \item Underneath the three resistances, a capacitor $C_l$ is added which represents the compliance of the lung tissue. Its charge represents the current lung volume.
    \item The visco-elastic structure of the lung tissue and chest wall is modelled by a Kelvin Body with resistance $R_{ve}$ and capacitance $C_{ve}$.
    \item The nonlinear chest wall compliance is modeled by a variable capacitance $C_{cw}$. The charge of this capacitor represents the current chestwall volume.
    \item The pressure applied by the muscles of the chest wall $P_{mus}$ is represented by a voltage source. This voltage source may also be used to model cardiac oscillations, although cardiac oscillations have a different origin.
    \item To ensure that the volume of the chest wall and lungs are both equal to the functional residual capacity (FRC) at the start of the simulation for a given positive end-expiratory pressure (PEEP), an extra voltage source $PipPEEP$ is added in series with a resistance $R_d$ to ensure this condition is met. The resistance $R_d$ is very high to ensure that there is no influence on the rest of the simulation.
\end{itemize}

Although the circuit may seem simple at first sight, the components in the circuit are not static scalars. 
To model the airways accurately, the components have complex nonlinear dynamics which are inter-coupled.
This makes an analytical solution not possible and simulation tools are needed to obtain a solution. The original model of Athanasiades et al. is implemented in the C programming language for 4 different test persons. We choose to implement it in a circuit simulator. Some minor modifications to the original model are made to improve speed and overcome the limitations of the circuit simulator without losing accuracy:
\begin{itemize}
    \item We replace the piece-wise continuous function that models the volume of the collapsible airways ($V_c$) by a function that has the same shape in the physiological region, but in contrast is completely continuous, and the first and second derivative are continuous:
    \begin{align}
        V_c = \frac{V_{cmax}}{((1+e^{-A_c(P_t-B_c)}))^{D_c}}.
    \end{align}
    $V_{cmax}$, $B_c$, $A_c$ and $D_c \in \mathbf{R}$ are patient-dependent parameters which determine the shape of the curve, and $P_t$ is the transmural pressure.
    This modification does not change the outcome of the model, but enables implementation of the model in a circuit simulator.
    The function for $R_c$ is also changed to a completely continuous function:
    \begin{align}
        R_c = K_c((1+e^{-A_c(P_t-B_c)}))^{2D_c},
    \end{align}
    with $K_c \in \mathbf{R}$.
    \item We replace the sine-function that is used to model the muscle pressure $P_{mus}$ in \cite{athanasiades2000energy} by a rounded trapezoid with a different slope for the rising edge and falling edge to model the muscle pressurization. This is more realistic and corresponds better to the shape of the muscle pressurization during ventilation \cite{vicario2015noninvasive, mojoli2016ventilator}.
    \item The order of the visco-elastic Kelvin body and the chest wall compliance are switched to enforce the condition: $V_{lung}+V_{collapsible}=V_{chestwall}$.
\end{itemize}
The remainder of the model for the healthy patient archetype is kept the same as the model by Athanasiades et al. \cite{athanasiades2000energy} and can be found in Table \ref{tab:patienthealthyeq} in Appendix \ref{app:lungequations}. For the disease archetypes, more modifications are made, which are addressed in the next section.

\subsection{Patient archetypes}
Athanasiades et al. \cite{athanasiades2000energy} estimate the parameters of the model based on measurements of four healthy test subjects. 
A synthetic data set suitable for machine learning has to be diverse and realistic. 
In clinical practice, different patient archetypes are present, which therefore should be part of the data set.
In particular, since the state of health of mechanically ventilated patients is usually compromised, it is important that the synthetic data set includes data of patients with different disease archetypes.
Therefore, from the literature in total four common disease types with different degrees of severity are selected that are known to change lung mechanics significantly: COPD with severity ``1'' and ``2'' (``2'' being the worst), obese ``1'' and ``2'', ARDS with severity ``1'', ``2'' and ``3'' and idiopathic fibrosis. In previous studies, a large effort is made to capture the change in mechanics for these different diseases using different types of measurement techniques.
The main features of these changes in mechanics can be incorporated into the proposed lung model by changing the lung model's parameters. The differences between the different patient archetypes are thus based on clinical measurements of mechanically ventilated patients.

\begin{figure}
    \centering
    \includegraphics[trim={0 1cm 0 1cm},clip,width=\columnwidth]{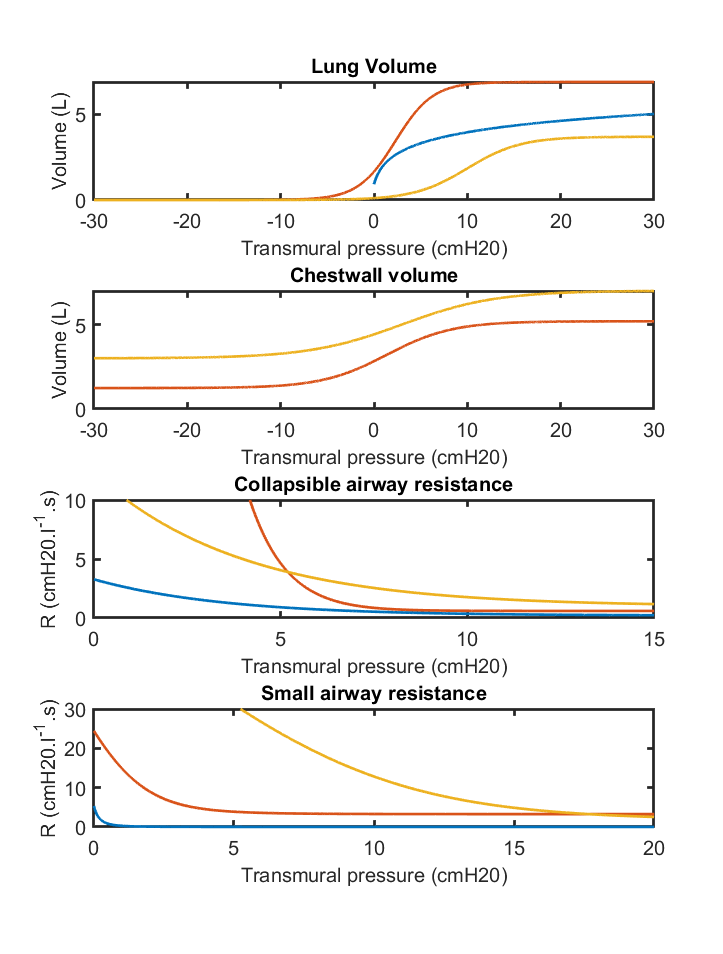}
    \caption{The pressure-volume and pressure-resistance curves for the healthy archetype (blue), the COPD archetype (red) and the ARDS archetype (yellow). Note that the different patient archetypes breathe in different ranges of transmural pressure.}
    \label{fig:ardspeep0}
\end{figure}

We propose the following changes based on experimental data from previous studies to model the patient archetypes:
\begin{itemize}
\item ARDS patients have a loss of oxygenated lung tissue caused by inflammation of the lungs, fluid accumulation in the lungs, and a partial collapse of the lung (atelectasis). For this reason, their lungs become less compliant and stiff \cite{pelosi2018close,kallet2003respiratory}. This results in a smaller residual volume, functional residual capacity, and total lung capacity.

ARDS is more complex and inhomogeneous than the model proposed in this paper can describe, moreover it differs from patient to patient \cite{sinha2019phenotypes}.
However, the model can capture the most important ARDS features when the lungs are open \cite{van2019recruitment}.

The smaller lung volume is modeled by changing the shape of the equation of the lung volume $V_l$. The original equation for $V_l$ described in \cite{athanasiades2000energy} is changed to the sigmoidal curve for ARDS described in Venegas et al. \cite{venegas1998comprehensive}, which is specifically developed for describing the lung volume of patients with ARDS:

\begin{align}
    V_l = \frac{A_l}{(1+e^{-B_l(P_t-
D_l)})}\,.
\label{eq:lungvolume}
\end{align}
$A_l$, $B_l$ and $D_l$ are constants determining the exact shape of the curve, and are shown in Table \ref{tab:patientchange} in Appendix \ref{app:paramvalue}. 
The change in the shape of the pressure-volume curve compared to the normal archetype is shown in Figure \ref{fig:ardspeep0}. 
Although this curve is used for the ARDS archetype, it can also be used for all other patient disease archetypes \cite{kondili2004pattern,ferreira2011sigmoidal}. Resistances of the airways are moderately increased compared to a healthy individual \cite{eissa1991analysis}, which is also shown in Figure \ref{fig:ardspeep0}. Finally, the time constant of the Kelvin body is increased to model the atelectasis and opening of the lungs.
These changes are sufficient to capture the most important features of ARDS ``1'', ``2'', and ``3''.
\item The term COPD is used for all patients with non-reversible obstructive airflow, however, there exists great variation between the different phenotypes of COPD \cite{miravitlles2012clinical}. For this work, we have made a selection of common changes that are found in COPD patients as compared to the healthy archetype. COPD is characterized by high airway resistance and low lung elastic recoil \cite{farre1998respiratory, guerin1993lung}. This results in high compliance of the lung tissue, which in turn results in higher lung volumes. For this reason, the total lung capacity, residual volume, and functional residual capacity are all higher than in a healthy person \cite{papandrinopoulou2012lung}. This is depicted in Figure \ref{fig:ardspeep0}. Note that COPD patients breathe in a lower range of transmural pressure than the normal archetype. The expiratory resistance is much higher in patients with COPD due to excessive central airway collapse especially during expiration \cite{murgu2006tracheobronchomalacia}.
The small airway resistance is also higher than the healthier patient archetype, because of the loss of elastic recoil, which normally opens the small airways. Late cycling is more prominent in the COPD archetype than in other archetypes.
\item In obesity, due to closure of the peripheral airways and due to the diaphragm being pressed towards the head in supine position \cite{parameswaran2006altered}, the total lung capacity and functional lung capacity are lower than in healthy subjects. The respiratory compliance is reduced, and therefore the lung volume, and the upper airway resistance is strongly increased mostly due to increased turbulence \cite{parameswaran2006altered, pelosi1997respiratory, zerah1993effects, brown2010reference}.

\item In idiopathic fibrosis, the lung tissue is very stiff and has low compliance. The residual volume, functional residual capacity, and total lung capacity are all lower compared to healthy individuals \cite{plantier2018physiology,marchioni2020ventilatory}.
The airway resistance is slightly lower than in healthy individuals. In patients with idiopathic fibrosis, the asynchrony ``early cycling" is more prevalent than in other patient archetypes.
\end{itemize}

All changes per patient archetype are shown in Table \ref{tab:patientchange} that we use in this paper. In clinical practice there exist more phenotypes of the same diseases, however, that is out of the scope of this work.
The model equations for the disease archetypes can be found in Table \ref{tab:patientdiseaseeq}. The corresponding parameters for each patient archetype can be found in Table \ref{tab:patientparam}.

\begin{table}[h!]
\centering
 \caption{\label{tab:patientchange} Overview of patient archetype changes compared to the healthy archetype. }
 \scalebox{0.7}{%
 \begin{tabular}{||c c c c c c||} 
 \hline
  & Healthy & Obese & ARDS & COPD & Idiopathic fibrosis \\ [0.5ex] 
 \hline\hline
 RV & - & $\downarrow \downarrow$ & $\downarrow$ & $\uparrow \uparrow$ &  $ \downarrow$ \\ 
 FRC & - & $\downarrow \downarrow$ & $\downarrow$ & $\uparrow \uparrow$ & $\downarrow \downarrow$ \\
 TLC & - & $\downarrow$ & $\downarrow$ & $\uparrow \uparrow$ & $\downarrow \downarrow$ \\
 Rup & - & $\uparrow \uparrow$ & $\uparrow$ & - & -\\
 Rinsp & - & $\uparrow$ & $\uparrow$ & $\uparrow$ & -\\
 Rexp & - & $\uparrow$ & $\uparrow$ & $\uparrow \uparrow$ & -\\
 Cl & - & $\downarrow$ & $\downarrow \downarrow$ & $\uparrow \uparrow$ & $\downarrow \downarrow$\\
 Ccw & - & $\downarrow$ & $\downarrow \downarrow$ & - & $\downarrow$\\
 Ctot & - & $\downarrow$ & $\downarrow \downarrow$ & $\uparrow \uparrow$ & $\downarrow$\\
 $\tau$ & - & $\uparrow \uparrow$ & $\downarrow$ & - & $\downarrow$\\ [1ex] 
 \hline
 \end{tabular}}
\end{table}

\subsection{Modelling of the breathing circuit and ventilator}
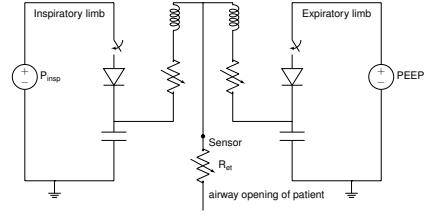
\begin{figure}
\begin{center}
\resizebox{0.75\columnwidth}{!}{%
\begin{circuitikz}[american voltages, transform shape]
\ctikzset{font=\fontfamily{phv}\selectfont}
\draw
(3,-2) to [vR, l_=R\textsubscript{et}] (3,-1)
(3,-2) to [short, l = airway opening of patient] (3,-3)
(3,-0.5) to [short] (3, 4)
(3,-0.5) to [short, *-, l^=Sensor] (3, -1)
(4, 4) to [L] (4,3)
(4,4) to [short] (3,4)
(2, 4) to [L] (2,3)
(2,4) to [short] (3,4)
(2,1) to [vR] (2,2)
(2,1) to [short] (2,0)
(2,0) to [short] (0,0)
(2,2) to [short] (2,3)
(0,0) to [C] (0,-1)
(0,1) to [short] (0,0)
(0,2) to [D] (0,1)
(0,3) to [short] (0,4)
(0,4) to [short, l = Inspiratory limb] (-3,4)
(0,3) to [switch] (0,2)
(-3,2) to [V, l = P\textsubscript{insp}] (-3,1)
(-3,2) to [short] (-3,4)
(-3,1) to [short] (-3,-2)
(0,-2) to [short] (-3,-2)
(0,-1) to [short] (0,-2)
(-2,-2) to (-2,-2) node[ground]{}
(4,0) to [short] (4,1)
(4,0) to [short] (6,0)
(4,1) to [vR] (4,2)
(4,2) to [short] (4,3)
(6,0) to [C] (6,-1)
(6,-2) to [short] (6,-1)
(6,4) to [short] (6,3)
(9,4) to [short, l = Expiratory limb] (6,4)
(6,1) to [short] (6,0)
(9,-2) to [short] (6,-2)
(6,2) to [D] (6,1)
(6,3) to [switch] (6,2)
(9,-2) to [short] (9,1)
(9,2) to [short] (9,4)
(9,2) to [V, l = PEEP] (9,1)
(8,-2) to (8,-2) node[ground]{};
\end{circuitikz}}
\end{center}
\caption{Equivalent circuit of the ventilator model. Note: the endotracheal tube and the resistances modeling the tubing system are not simple resistances, but are modeled by Rohrer's equation and depend on the flow through the elements. A control algorithm determines when the switches of the ventilator are opened and closed.}
    \label{fig:ventilatorcircuit}
\end{figure}

To model patient-ventilator interactions, the adapted lung model by Athanasiades et al. \cite{athanasiades2000energy} is extended with the simple ventilator circuit shown in Figure \ref{fig:ventilatorcircuit}.
The model consists of the following parts:
\begin{itemize}
    \item The endotracheal tube $R_{et}$ (the tube which is inserted in the trachea through the mouth), is modeled as a variable resistance to include turbulent airflow. The endotracheal tube adds an extra resistance $R_{et}$ in series with the airway resistances from the patient. We choose the same approach as Flevari et al. \cite{flevari2011rohrer}, where the resistance of the endotracheal tube is modeled by Rohrer's equation which is a model for turbulence.
    Moreover, Flevari et al. determine the equation parameters for endotracheal tubes with different diameters. In our simulation setup, we can easily change between different parameters and therefore simulate different endotracheal tubes.
   \item The tubing system connects the endotracheal tube and the ventilator. This system is split into an inspiratory circuit and an expiratory circuit.
   Both are modeled by an inductor (inertance), capacitor (compliance), and resistor, see Figure \ref{fig:ventilatorcircuit}.
   Wenzel et al. \cite{wenzel2017coaxial} model the resistance of breathing circuits with Rohrer's equation for different brands and types of tubes. This equation takes into account the turbulence in the tubes, which is important and should not be neglected; therefore, we take the same approach. The inertance and compliance of these breathing tubes are also measured in Wenzel et al. \cite{wenzel2017coaxial} and are also included in the model.
   \item The ventilator itself is modeled by two pressure sources, different unidirectional valves, and parasitic elements. The pressure sources in the ventilator determine the positive end-expiratory pressure (PEEP), and the peak inspiratory pressure ($P_{insp}$). These are the pressures at which the ventilator operates.
   The ventilator pressurization is modeled by two voltage sources that represent the pressure sources.
   The valves are modeled by switches and by unidirectional diode-like elements, which block inspiration from the contaminated expiratory connection.
   Valves are opened and closed when the ventilator is triggered and when it cycles; however, this is not drawn in Figure \ref{fig:ventilatorcircuit}. This results in a block wave or a trapezoidal wave for the ventilator.
   \end{itemize}
The pressure and the flow in ventilated patients are usually measured at the beginning of the endotracheal tube.


\subsection{Modelling cardiac oscillations and measurement noise}
The model generates noise-free measurements.
For machine learning data it is important to have simulations that have the same characteristics as experimental data.
For this reason, we add filtered white noise with a bandwidth of 0-15 Hz to the pressure and flow curves. The noise amplitude is chosen in such a way that it mimics the noise in experimental waves. Another artifact that is often observed in ventilator waveforms originates from cardiac oscillations. These are oscillations caused by the pulmonary blood change due to the pulsating heart \cite{arbour2009cardiogenic}. We model this as a sine wave that is added to the muscle pressure.
\subsection{Automatic labeling and identifying asynchronies}\label{sec:async}
Since both ventilator and patient are fully controlled by the model, the data can be labeled automatically. Every regular breath with correct pressurization consists, next to the pressure, flow, and volume waveforms, of four time labels which are defined as follows:
\begin{itemize}
    \item The start of patient inspiration: We define this as the start of the patient's muscle contraction. This is the same definition as often employed in the literature, e.g., in \cite{chao1997patient} and \cite{fabry1995analysis}.
    \item The end of patient inspiration. We define this as the moment when the patient's muscle effort becomes zero again. This is the same definition as used in Fabry et al. \cite{fabry1995analysis}, but note that it differs from the definition used in Mojoli et al. \cite{mojoli2016ventilator}, who use half relaxation as the optimal cycling/end of inspiration point.
    \item The start of the ventilator pressurization (triggering): the moment the ventilator switches to inspiration.
    \item The end of ventilator pressurization (cycling): The moment the ventilator switches to expiration.
\end{itemize}

To determine whether asynchronies are present, we classify breaths by looking at the difference between the start of patient inspiration effort and the start of ventilator pressurization (the start-inspiration delay or inspiratory response delay \cite{fabry1995analysis}, to look for trigger asynchronies) and the difference between the end of patient inspiration effort and the end of ventilator pressurization (the end-inspiration delay or expiratory response delay \cite{fabry1995analysis}, to look for cycling asynchronies).

It should be noted that the inspiratory response delay increases when the inspiratory trigger is less sensitive, when there is a large tidal volume or when expiratory gas flow is restricted  \cite{fabry1995analysis}.
The expiratory response delay is increased due to lower inspiratory peak flow, a higher level of pressure support, or less sensitive cycling criteria \cite{fabry1995analysis}.

To classify whether a breath is an asynchrony or a regular breath, these delays need to fall into a certain margin.
We employ the same margins as Bakkes et al. \cite{bakkes2020machine} since the machine learning algorithm that we use for validation uses the same criteria: 
A normal breath has an end-inspiration delay larger than -100 ms and smaller than 300 ms. The start-inspiration delay must be lower than 250 ms. All other breaths are asynchronies.

Although there are more asynchronies described in the literature, for this study we focus exclusively on the most common asynchronies during PSV and use the following criteria to classify these asynchronies:

\begin{itemize}
    \item Early cycling: the duration of ventilator pressurization is shorter than patient inspiratory effort. More specifically, we define that the end-inspiration delay must be shorter than -100 ms.
    \item Late cycling: the duration of the ventilator pressurization is longer than the patient inspiratory effort. The end-inspiration delay is longer than 300 ms.
    \item Delayed inspiration: There is a significant trigger delay between the patient inspiration and the ventilator inspiration; the ventilator inspiration starts late compared to the patient inspiration. The start-inspiration delay is longer than 250 ms.
    \item Ineffective effort is an exception to the asynchronies above: patient effort is not followed by a ventilator pressurization. In other words, there is a patient effort but the ventilator is not triggered, the start-inspiration delay and end-inspiration delay are therefore not defined.
\end{itemize}

\begin{figure}
    \centering
    \includegraphics[width=\columnwidth]{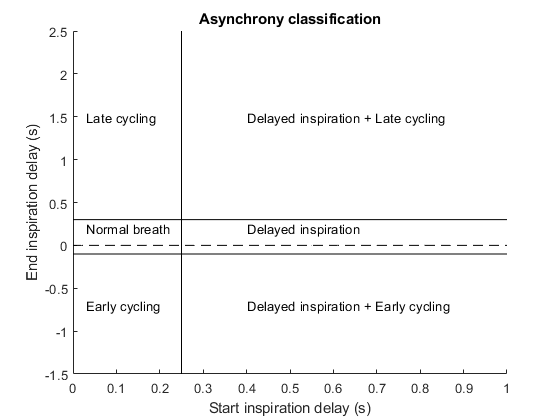}
    \caption{The classification system used for asynchronies in this work. The classification is the same as in Bakkes et al. \cite{bakkes2020machine}. The x-axis is the start inspiration delay, on the y-axis the end inspiration delay is shown.}
    \label{fig:asynchronyclass}
\end{figure}

The margins are visualized in Figure \ref{fig:asynchronyclass}. However, since the machine learning algorithm estimates the start inspiration delay and the end inspiration delay, it is possible to tune the margins. This is however out of the scope of this paper.

The asynchronies are generated in a random sequence throughout the synthetic data, taking into account that some types of asynchronies are more prominent in particular patient archetypes. 

\subsection{Implementation and generation of the data set} \label{sec:generation}
The lung and ventilator models are implemented in LT Spice XVII \cite{Nagel:M382}, which is an analog electronic circuit simulator. A wrapper in MATLAB R2019b \cite{MATLAB:2019} is written in order to facilitate rapid changing of the parameters, improved plotting functionality, and automated runs. 
The input muscle waveform $P_{mus}$, and the ventilator triggering and cycling points are also generated by MATLAB R2019b.
The MATLAB wrapper consists of three stages: The first stage is when an LT Spice run is performed with only PEEP. This is done to calculate the triggering times. In the next run, the ventilator is triggered at the correct times and is kept on until a certain time limit is reached. With this data, the correct cycling times are calculated. The last run is done with both correct triggering times and cycling times.
The resulting timing points are saved together with the generated waveforms. 
The simulation is $\sim$70x faster than real-time on an Intel Core i7 7th Gen.

In order to check the feasibility of the model and generated data, we generate a data set with the model described above. 
Before each simulation, the initialization value for the interpleural pressure for the correct PEEP ($PipPEEP$, see Figure \ref{fig:lungmodel}) needs to be calculated in such a way that the chest wall volume and lung volume are equal.

We generate the simulations for the five different patient archetypes and we distribute the asynchronies randomly throughout the dataset.
We take into account that some asynchronies are more prominent in particular disease archetypes, we, therefore, specify a different distribution of asynchronies for each disease archetype.
 
We generate simulations that are always 120 seconds long and contain 30-40 breaths. Longer simulations are possible but this may lead to numerical errors in some cases.
The PEEP and inspiratory pressure were chosen in such a way that the tidal volume was always set to around 500~mL. This corresponds to 7~mL/kg with the assumption of an average patient being 70~kg \cite{fabry1995analysis} \cite{zhang2020detection}.

\subsection{Validation methods}
 We are particularly interested in three aspects of the model: whether the main features of the waveforms are modeled correctly, whether the asynchronies are modeled correctly, and whether the different patient archetypes are modeled correctly.
To evaluate the model and the data it generates, we propose three methods to check the validity of the data.
\begin{itemize}
    \item The synthetic waveforms are compared to clinical ventilator waveforms found in literature \cite{colombo2011efficacy, gholami2018replicating,chao1997patient, thille2006patient, georgopoulos2006bedside, dres2016monitoring} by visual inspection. A clinical expert (AdB) has reviewed and commented on the synthetic waveforms. The review is used to check whether there are visual inconsistencies in the data.
    \item  Machine learning trained on validated clinical data is tested on the synthetic data. We used the algorithm proposed by Bakkes et al. \cite{bakkes2020machine}, which is trained on clinical data. The data were obtained at Fondazione I.R.C.C.S. Policlinico San Matteo (Pavia, Italy), and contains 4275~breaths from 15 patients, who were not able to breathe independently. The data was labeled by a clinical expert who indicated the inspiratory and expiratory efforts of the patients.
    The algorithm is trained to recognize the time points described in Section \ref{sec:async} in unlabeled data.
    The difference in results between the algorithm tested on the clinical and synthetic data might indicate that there is a disagreement between the clinical data and the synthetic data.
\end{itemize}

\section{Results} 

\label{sec:results}
\subsection{Comparison with clinical data}
\begin{figure*}[ht]
    \centering
    \includegraphics[trim={1cm 0 3.5cm 1cm},clip,width=\textwidth]{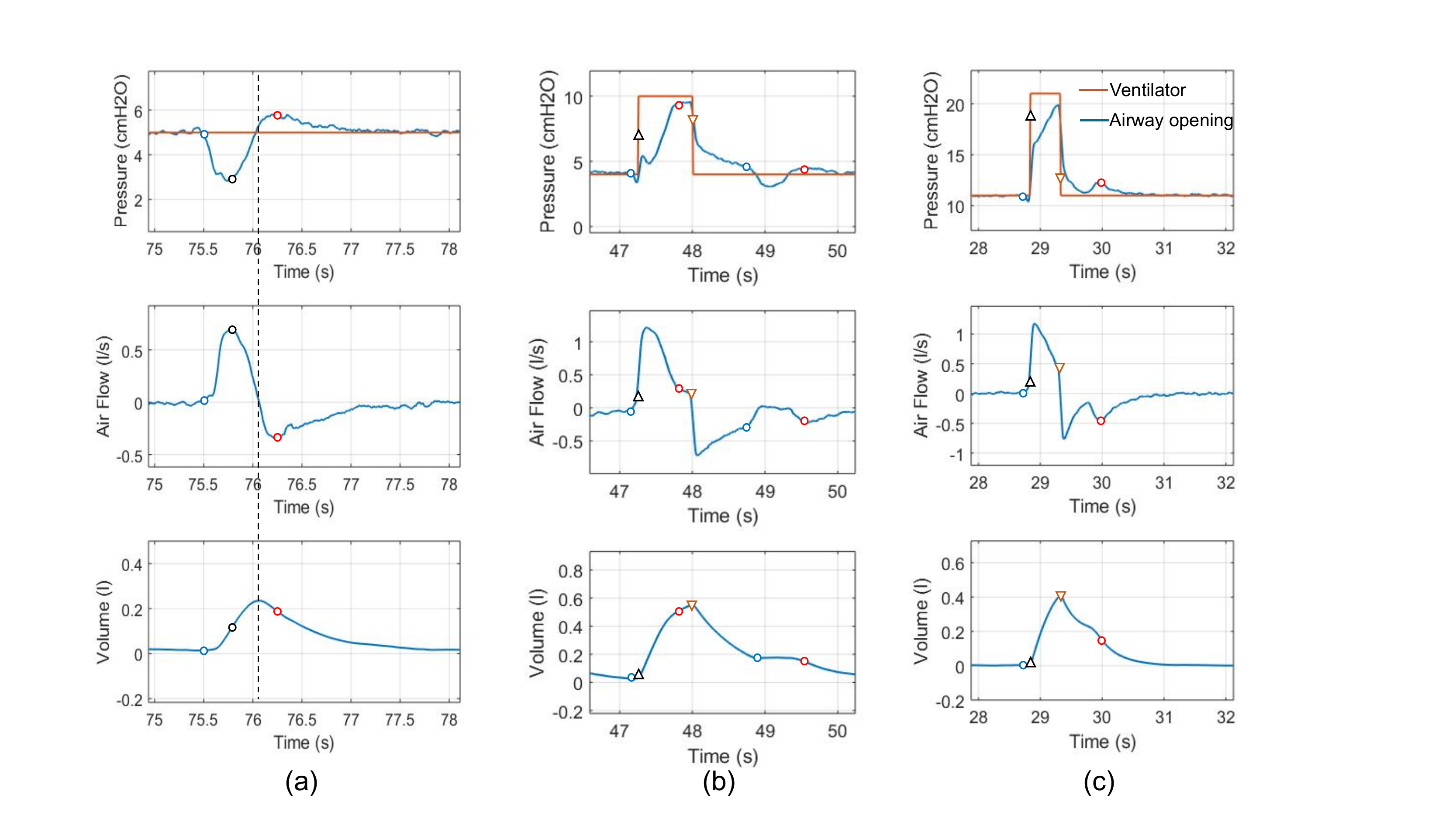}
    \caption{Simulations showing the different time points in the simulated waveforms, and the different ventilator settings. (a) shows the pressure, flow and volume for a ventilator with CPAP (only PEEP). The blue, black and red markers show the start of patient inspiration, maximum effort and end of patient effort respectively. (b) shows a late cycling breath and an ineffective effort. The triangular markers show the start of the ventilator trigger and the cycling of the ventilator. The black circle marker is no longer shown here for clarity. (c) The same waveform is shown but now with a ventilator pressurization with a higher PEEP and $P_{insp}$. The end of patient inspiration (red round marker) is later than the end of the ventilator pressurization (yellow triangle marker) this breath is therefore early cycling.}
    \label{fig:timepoints}
\end{figure*}

\begin{figure*}[ht]
    \centering
   \includegraphics[width=\textwidth,]{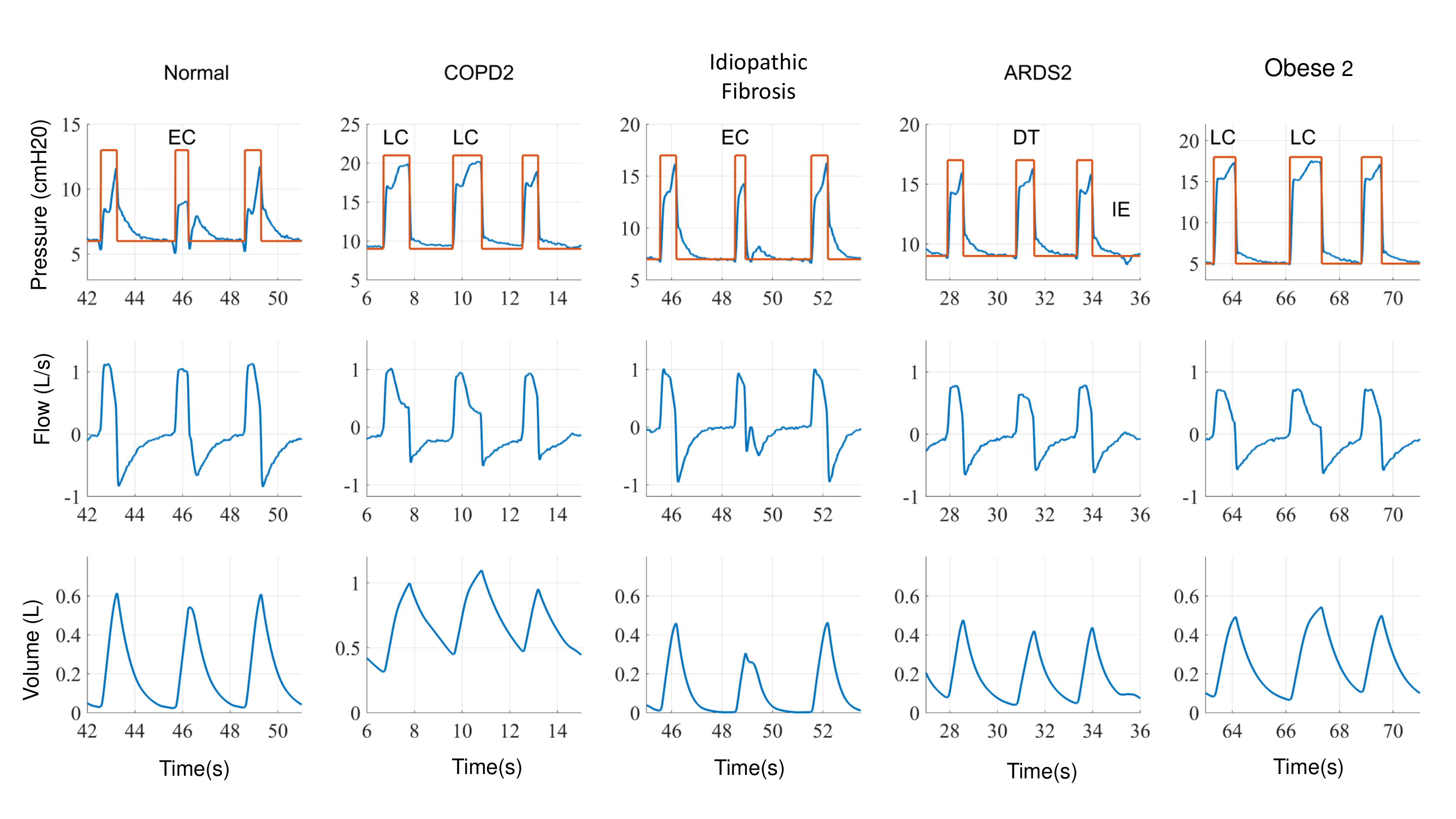}
   \caption{The subfigures depict pressure, flow and volume waveforms of 5 different patient archetypes. In the normal archetype, the middle breath is early cycling. In the COPD2 archetype, two times late cycling is shown. In the idiopathic fibrosis archetype, early cycling is shown. The ARDS2 archetype shows delayed triggering and an ineffective effort while the OBESE2-archetype shows two late cyclings.}
   \label{fig:simulationexamples}
\end{figure*}

In Figure \ref{fig:timepoints} three simulated breaths are shown. The markers in the ﬁgure indicate the automatically generated annotations. Figure \ref{fig:timepoints}a shows a breath with patient effort using only PEEP (the ventilator is not triggered). This is the situation during continuous positive airway pressure (CPAP) or during an ineffective effort. The start of inspiration, maximum effort, and end-of-effort markers are indicated. The expiratory phase starts when flow changes sign (dotted line). Figure 4b shows the simulated waves for a late cycling event and an ineffective effort during expiration. In the late cycled breath, both the ventilation trigger and cycling markers, along with the start and end of patient effort markers, are shown. The shape of both pressure and flow rate shows the characteristic features of a late cycling event. The pressure at the airway opening is initially lower than the ventilator inspiratory pressure; the difference depends on the inspiratory tube resistance, airway resistance, and airway flow rate.  During active inspiration, the flow wave has a non-exponential shape. At the end of patient inspiratory effort, the flow wave changes into an exponential shape. This difference in flow wave shape is an important marker for the end of patient effort and this specific time can be extracted from the flow wave. It coincides with the end-of muscle effort time. During the expiration phase, both pressure and flow waves decrease exponentially. Note that in this figure an ineffective effort occurs during expiration. Both pressure and flow waves show the characteristic shape and features of such an event. A drop in pressure and a decrease in flow are observed. The flow rate is limited by the unidirectional valve in the expiratory tube.
Figure \ref{fig:timepoints}c shows an early cycling asynchrony. The time point when patient inspiration ends lies after the time point when the ventilator cycles, resulting in a pattern in pressure and flow waves which is characteristic for early cycling. The minima in pressure and maxima inflow after cycling can be observed when cycling occurs between the start of inspiration and maximum inspiratory effort. The start of exponential pressure and flow decay occurs at the end of the inspiration effort. 
In Figure \ref{fig:simulationexamples} the simulation results for pressure, flow, and tidal volume for five patient archetypes are shown. The PEEP, maximum inspiratory pressure $P_{insp}$, and duration of the inspiratory and expiratory times are chosen to be compliant with a lung-protective ventilation scheme of 500~mL tidal volume (corresponding to a 70~kg adult male). Three breaths are shown for each archetype, a combination of normal and asynchronous ventilations. For the COPD case, two late cycling events are shown; for the fibrosis archetype, an early cycling event is shown. A delayed trigger and an ineffective effort are shown for the ARDS archetype. For the obese two archetypes with two late cycling events are shown. 

\subsection{Machine learning results}
\begin{figure*}[ht]
    \centering
    \includegraphics[width=\textwidth]{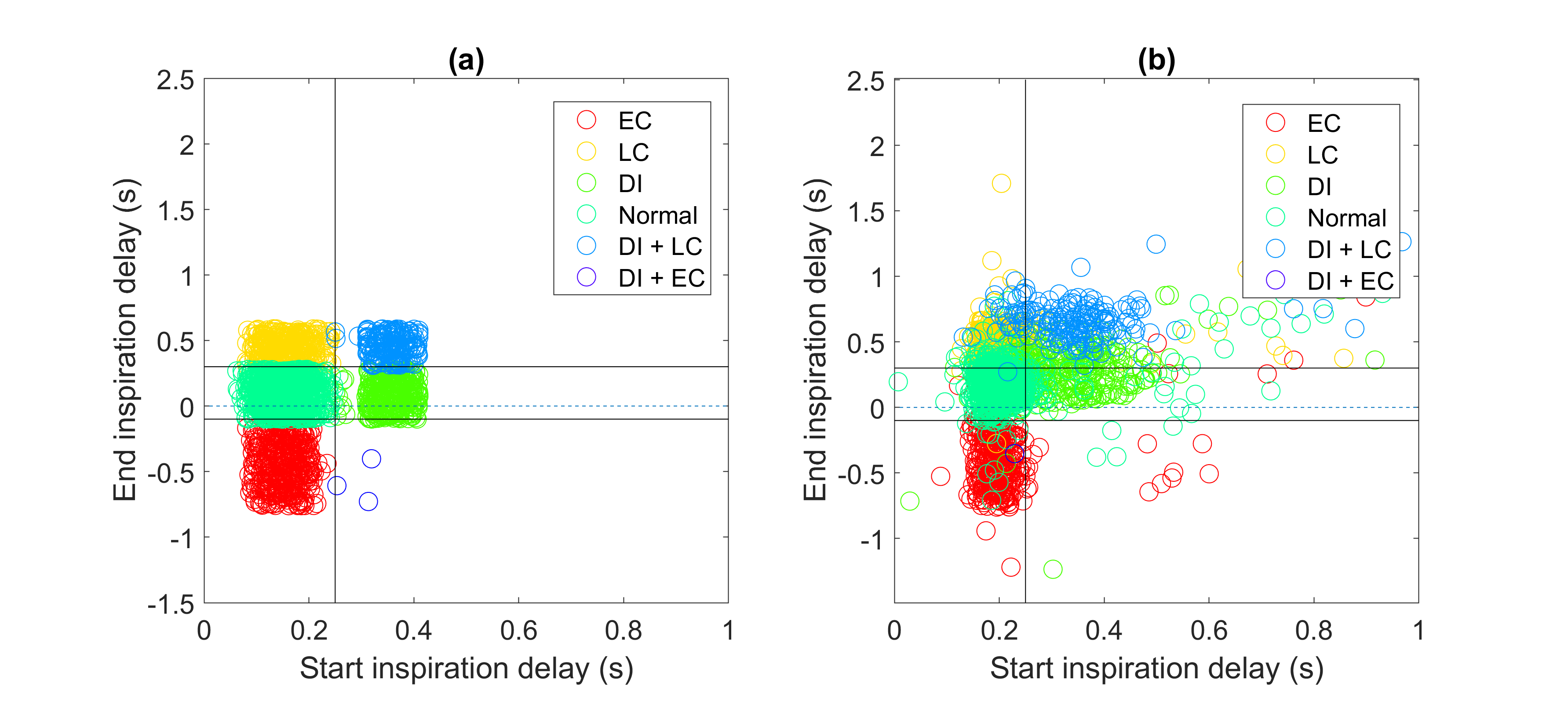}
    \caption{(a) This picture shows the true inspiration delays. (b) This picture shows the estimated inspiration delays by the machine learning algorithm trained on clinical data. Start-inspiration delay is the delay between patient inspiration and triggering of the ventilator, end-inspiration delay is the difference between the start of patient expiration and triggering of the ventilator. No points for ineffective effort (IE) are shown, since the ventilator is not triggered and hence no delays can be calculated.}
    \label{fig:tomsresults}
\end{figure*}

\begin{figure*}[ht]
    \centering
    \includegraphics[width=\textwidth]{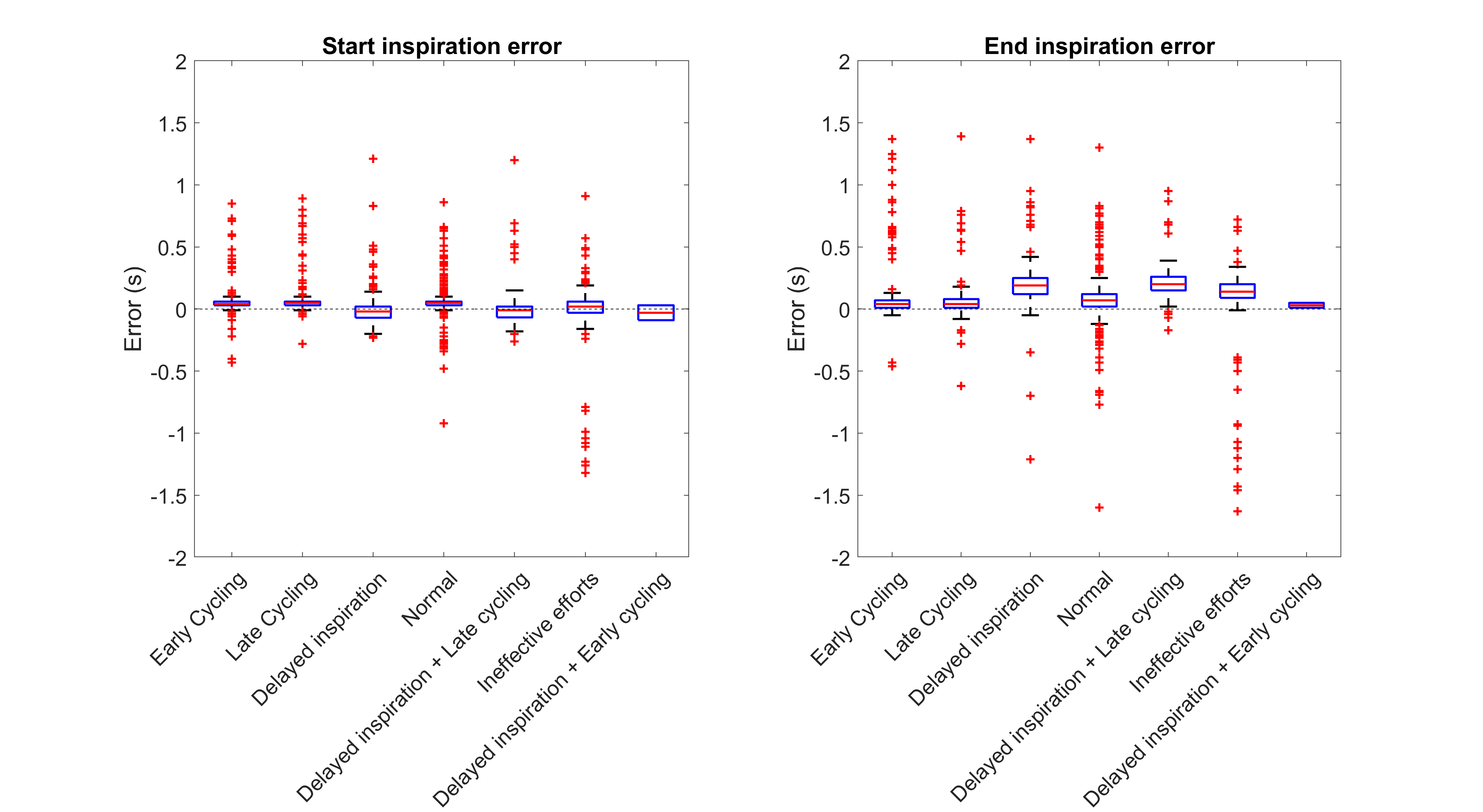}
    \caption{On the left the error between the true start-inspiration error and the estimated start-inspiration error, on the left the error between the true end-inspiration error and the estimated end-inspiration error by the machine learning algorithm.}
    \label{fig:tomsresults_boxplot}
\end{figure*}

Figure \ref{fig:tomsresults}a shows a plot for 2000 breaths of the true start-inspiration delay and end-inspiration delay (as deﬁned in Section \ref{sec:async}). This shows the distributions of asynchronies in the simulated data set. Figure \ref{fig:tomsresults}b shows the estimated start-inspiration delay and end-inspiration delay as estimated by the machine learning algorithm. The difference between the ground truth simulations and estimated machine learning results is shown in Figure \ref{fig:tomsresults_boxplot}. The median error for the majority of the breaths is close to zero and the interquartile range is small; however, some outliers are observed. Note that the number of outliers is much smaller than the correctly estimated times within the 25-75\% range. The median value error for the inspiration error is small for all breath types. However, for the end-inspiration error, there are systematic deviations of the median error value for the delayed inspiration (DI) types and the median error is around +200 ms.  
Table \ref{tab:perfmet} shows the performance metrics per asynchrony class. It shows a lower true positive rate for delayed inspiration and a lower positive predictive value for delayed inspiration + late cycling breaths. This corresponds to the observation in the previous figures.


\label{app:confmat}
\begin{table*}[hbt!]
\centering
 \caption{\label{tab:perfmet} Performance metrics}
 \scalebox{0.9}{%
 \begin{tabular}{||c c c c c c c||} 
 \hline
  & EC & LC & DI & Normal & DI LC & DI EC \\ [0.5ex] 
 \hline\hline
True positive rate & 0.86 & 0.88 & 0.43 & 0.78 & 0.82 & 0\\ 
True negative rate & 0.99 & 0.91 & 0.98 & 0.96 & 0.91 & 0.99\\
Positive predictive value & 0.94 & 0.64 & 0.76 & 0.94 & 0.46 & 0\\
Balanced Accuracy & 0.93 & 0.89 & 0.71 & 0.87 &0.86 &0.50\\
 \hline
 \end{tabular}}
\end{table*}

\section{Discussion} \label{sec:discussion}
Due to the emerging need for large, labeled data sets for training and testing of machine learning for mechanical ventilation, this paper explored generating a synthetic dataset with automatic labeling of ventilator waveforms including different types of asynchronies.
The method used in this paper combined an already available nonlinear lung model with a simple ventilator model. Some small adjustments were made to generate synthetic ventilator waveforms from this combined lung-ventilator model using a circuit simulator.
Both the comparison with clinical data by an expert and the machine learning results suggests that it is possible to generate an automatically labeled synthetic data set with the most important features of a clinical dataset.

The evaluation of the clinical expert shows that the waveforms are recognizable as patient-ventilator waveforms. However, there are still differences between experimental waves and the simulated waves presented in this paper.
The differences are most likely caused by simplifications in the lung model and the ventilator model. Clinical mechanical ventilators have more complex control algorithms that depend on the type and brand of the ventilator \cite{richard2002bench}.
The ventilator model in this work is not complex enough to model these differences and requires three runs to simulate simple control loops, which leads to artifacts in the waveforms. The lung model is also less complex than a patient. 

The main finding of the machine learning results is that there is an overestimation of the median end-inspiration error (Figure \ref{fig:tomsresults_boxplot}b) for all asynchrony types.
Still, for most asynchrony classes the machine learning algorithm has a balanced accuracy higher than 0.85. However, especially the delayed inspiration asynchrony suffers from a lower accuracy.
This is also to some extend visible in the results on the experimental data set in Bakkes et al. \cite{bakkes2020machine}, although it is more pronounced in the synthetic data. This was caused by the method used to calculate the triggering time, which required three runs, and did not take the effect from adjacent breaths into account. At this moment this is fixed.
Another factor that caused this was the patient effort which was manipulated for delayed inspiration to obtain a delayed trigger. For further work, the muscle waveform should still be refined.
The combinations of asynchronies and patient archetypes were chosen in such a way that they corresponded to experimental data, but discrepancies might still be present. There might also be more variation present in the experimental data set.
The procedure of generating the data set from 3 runs is different from the clinical case, this may cause some interactions between adjacent breaths even when pressure triggering and flow cycling is used.

The results do fit in the theory that machine learning algorithms might be trained on the synthetic dataset.
Moreover, generating the synthetic dataset has given us new insight into the mechanisms of asynchronies, and might be useful for educational purposes.

We want to emphasize that the model used in this paper is a simplification of the real interaction between the patient and the ventilator. The model omits the reaction of the patient on being mechanical ventilation \cite{kondili2003patient}. Also, longer-term effects of mechanical ventilation are not modeled. Ideally, the parameters in Table \ref{app:paramvalue} would change over time, depending on what would be happening to the patient.
Whether this leads to problems with using this data set for machine learning, still needs to be investigated.

The methodology of checking the synthetic data by visual inspection and by machine learning is limited. It is sometimes difficult to judge whether a certain observation in the machine learning results is caused by the data or by the algorithm itself. For more research on the similarities between the clinical dataset and the experimental dataset, more similarity metrics could be included in future research. However, we did not aim to recreate the clinical data exactly.
\section{Conclusion}
\label{sec:conclusion}
This study demonstrates how an accurate labeled synthetic data set of patient-ventilator waveforms can be generated for training and testing machine learning algorithms to detect patient-ventilator asynchronies.
The patient-ventilator model used in this paper is, however, still a simplification, which might introduce certain artifacts in the synthetic waveforms and makes it impossible to incorporate certain effects in the data. Future research will focus on testing the effect of synthetic data on the training and testing phase of machine learning algorithms.

\section{Acknowledgements}
The authors would like to thank Professor Francesco Mojoli.

This research did not receive any specific grant from funding agencies in the public, commercial, or not-for-profit sectors.

All the authors have no conflict of interest to declare.
\printbibliography

@article{arnal2018parameters,
  title={Parameters for simulation of adult subjects during mechanical ventilation},
  author={Arnal, Jean-Michel and Garnero, Aude and Saoli, Mathieu and Chatburn, Robert L},
  journal={Respiratory care},
  volume={63},
  number={2},
  pages={158--168},
  year={2018},
  publisher={Respiratory Care}
}

@article{venegas1998comprehensive,
  title={A comprehensive equation for the pulmonary pressure-volume curve},
  author={Venegas, Jos{\'e} G and Harris, R Scott and Simon, Brett A},
  journal={Journal of Applied Physiology},
  volume={84},
  number={1},
  pages={389--395},
  year={1998},
  publisher={American Physiological Society Bethesda, MD}
}

@article{kaczka2001inspiratory,
  title={Inspiratory lung impedance in COPD: effects of PEEP and immediate impact of lung volume reduction surgery},
  author={Kaczka, David W and Ingenito, Edward P and Body, Simon C and Duffy, Sabine E and Mentzer, Steven J and DeCamp, Malcolm M and Lutchen, Kenneth R},
  journal={Journal of Applied Physiology},
  volume={90},
  number={5},
  pages={1833--1841},
  year={2001},
  publisher={American Physiological Society Bethesda, MD}
}

@article{papandrinopoulou2012lung,
  title={Lung compliance and chronic obstructive pulmonary disease},
  author={Papandrinopoulou, D and Tzouda, V and Tsoukalas, G},
  journal={Pulmonary medicine},
  volume={2012},
  year={2012},
  publisher={Hindawi}
}

@article{athanasiades2000energy,
  title={Energy analysis of a nonlinear model of the normal human lung},
  author={Athanasiades, A and Ghorbel, F and Clark Jr, JW and Niranjan, SC and Olansen, J and Zwischenberger, JB and Bidani, A},
  journal={Journal of Biological Systems},
  volume={8},
  number={02},
  pages={115--139},
  year={2000},
  publisher={World Scientific}
}

@article{gholami2018replicating,
  title={Replicating human expertise of mechanical ventilation waveform analysis in detecting patient-ventilator cycling asynchrony using machine learning},
  author={Gholami, Behnood and Phan, Timothy S and Haddad, Wassim M and Cason, Andrew and Mullis, Jerry and Price, Levi and Bailey, James M},
  journal={Computers in biology and medicine},
  volume={97},
  pages={137--144},
  year={2018},
  publisher={Elsevier}
}

@article{de2019patient,
  title={Patient-ventilator asynchronies during mechanical ventilation: current knowledge and research priorities},
  author={de Haro, Candelaria and Ochagavia, Ana and L{\'o}pez-Aguilar, Josefina and Fernandez-Gonzalo, Sol and Navarra-Ventura, Guillem and Magrans, Rudys and Montany{\`a}, Jaume and Blanch, Llu{\'\i}s and others},
  journal={Intensive care medicine experimental},
  volume={7},
  number={1},
  pages={43},
  year={2019},
  publisher={Springer}
}

@article{slutsky_ventilator-induced_2013,
	title = {Ventilator-induced lung injury},
	volume = {369},
	issn = {1533-4406},
	doi = {10.1056/NEJMra1208707},
	language = {eng},
	number = {22},
	journal = {The New England Journal of Medicine},
	author = {Slutsky, Arthur S. and Ranieri, V. Marco},
	month = nov,
	year = {2013},
	pmid = {24283226},
	pages = {2126--2136}
}

@article{blanch2015asynchronies,
  title={Asynchronies during mechanical ventilation are associated with mortality},
  author={Blanch, Llu{\' i}s and Villagra, Ana and Sales, Bernat and Montanya, Jaume and Lucangelo, Umberto and Luj{\'a}n, Manel and Garc{\'\i}a-Esquirol, Oscar and Chac{\'o}n, Encarna and Estruga, Anna and Oliva, Joan C and others},
  journal={Intensive care medicine},
  volume={41},
  number={4},
  pages={633--641},
  year={2015},
  publisher={Springer}
}

@article{miravitlles2012clinical,
  title={Clinical phenotypes of COPD: identification, definition and implications for guidelines},
  author={Miravitlles, Marc and Calle, Myriam and Soler-Catalu{\~n}a, Juan Jos{\'e}},
  journal={Archivos de Bronconeumolog{\' i}a (English Edition)},
  volume={48},
  number={3},
  pages={86--98},
  year={2012},
  publisher={Elsevier}
}

@article{kondili2003patient,
  title={Patient--ventilator interaction},
  author={Kondili, E and Prinianakis, G and Georgopoulos, D},
  journal={British Journal of Anaesthesia},
  volume={91},
  number={1},
  pages={106--119},
  year={2003},
  publisher={Oxford University Press}
}

@article{van2019recruitment,
  title={Recruitment maneuvers and higher PEEP, the so-called open lung concept, in patients with ARDS},
  author={van der Zee, Philip and Gommers, Diederik},
  journal={Critical Care},
  volume={23},
  number={1},
  pages={1--7},
  year={2019},
  publisher={BioMed Central}
}

@misc{gattinoni2020covid,
  title={COVID-19 pneumonia: different respiratory treatments for different phenotypes?},
  author={Gattinoni, Luciano and Chiumello, Davide and Caironi, Pietro and Busana, Mattia and Romitti, Federica and Brazzi, Luca and Camporota, Luigi},
  year={2020},
  publisher={Springer}
}

@article{sinha2019phenotypes,
  title={Phenotypes in ARDS: Moving Towards Precision Medicine},
  author={Sinha, Pratik and Calfee, Carolyn S},
  journal={Current opinion in critical care},
  volume={25},
  number={1},
  pages={12},
  year={2019},
  publisher={NIH Public Access}
}

@article{thille2006patient,
  title={Patient-ventilator asynchrony during assisted mechanical ventilation},
  author={Thille, Arnaud W and Rodriguez, Pablo and Cabello, Belen and Lellouche, Fran{\c{c}}ois and Brochard, Laurent},
  journal={Intensive care medicine},
  volume={32},
  number={10},
  pages={1515--1522},
  year={2006},
  publisher={Springer}
}

@article{subira2018minimizing,
  title={Minimizing asynchronies in mechanical ventilation: current and future trends},
  author={Subir{\`a}, Carles and de Haro, Candelaria and Magrans, Rudys and Fern{\'a}ndez, Rafael and Blanch, Llu{\'\i}s},
  journal={Respiratory care},
  volume={63},
  number={4},
  pages={464--478},
  year={2018},
  publisher={Respiratory Care}
}

@article{loo2018machine,
  title={A machine learning model for real-time asynchronous breathing monitoring},
  author={Loo, NL and Chiew, YS and Tan, CP and Arunachalam, G and Ralib, AM and Mat-Nor, M-B},
  journal={IFAC-PapersOnLine},
  volume={51},
  number={27},
  pages={378--383},
  year={2018},
  publisher={Elsevier}
}

@inproceedings{gaidon2016virtual,
  title={Virtual worlds as proxy for multi-object tracking analysis},
  author={Gaidon, Adrien and Wang, Qiao and Cabon, Yohann and Vig, Eleonora},
  booktitle={Proceedings of the IEEE conference on computer vision and pattern recognition},
  pages={4340--4349},
  year={2016}
}

@article{liu1998airway,
  title={Airway mechanics, gas exchange, and blood flow in a nonlinear model of the normal human lung},
  author={Liu, CH and Niranjan, SC and Clark Jr, JW and San, KY and Zwischenberger, JB and Bidani, A and (With the Technical Assistance of HB Winnike, C. Vanouye, and JB Olansen)},
  journal={Journal of applied physiology},
  volume={84},
  number={4},
  pages={1447--1469},
  year={1998},
  publisher={American Physiological Society Bethesda, MD}
}

@inbook{bates_2009, place={Cambridge}, title={The linear single-compartment model}, DOI={10.1017/CBO9780511627156.004}, booktitle={Lung Mechanics: An Inverse Modeling Approach}, publisher={Cambridge University Press}, author={Bates, Jason H. T.}, year={2009}, pages={37–61}}

@article{flevari2011rohrer,
  title={Rohrer's constant, K2, as a factor of determining inspiratory resistance of common adult endotracheal tubes},
  author={Flevari, AG and Maniatis, N and Kremiotis, TE and Siempos, I and Betrosian, AP and Roussos, C and Douzinas, E and Armaganidis, A},
  journal={Anaesthesia and intensive care},
  volume={39},
  number={3},
  pages={410--417},
  year={2011},
  publisher={SAGE Publications Sage UK: London, England}
}

@article{wenzel2017coaxial,
  title={Coaxial tubing systems increase artificial airway resistance and work of breathing},
  author={Wenzel, Christin and Schumann, Stefan and Spaeth, Johannes},
  journal={Respiratory care},
  volume={62},
  number={9},
  pages={1171--1177},
  year={2017},
  publisher={Respiratory Care}
}

@inproceedings{eggert2015benefit,
  title={On the benefit of synthetic data for company logo detection},
  author={Eggert, Christian and Winschel, Anton and Lienhart, Rainer},
  booktitle={Proceedings of the 23rd ACM international conference on Multimedia},
  pages={1283--1286},
  year={2015}
}

@inproceedings{abt2014plea,
  title={A plea for utilising synthetic data when performing machine learning based cyber-security experiments},
  author={Abt, Sebastian and Baier, Harald},
  booktitle={Proceedings of the 2014 Workshop on Artificial Intelligent and Security Workshop},
  pages={37--45},
  year={2014}
}

@article{ghorbani2019dermgan,
  title={DermGAN: Synthetic Generation of Clinical Skin Images with Pathology},
  author={Ghorbani, Amirata and Natarajan, Vivek and Coz, David and Liu, Yuan},
  journal={arXiv preprint arXiv:1911.08716},
  year={2019}
}

@article{pelosi1997respiratory,
  title={Respiratory system mechanics in sedated, paralyzed, morbidly obese patients},
  author={Pelosi, P and Croci, M and Ravagnan, I and Cerisara, M and Vicardi, P and Lissoni, A and Gattinoni, L},
  journal={Journal of Applied Physiology},
  volume={82},
  number={3},
  pages={811--818},
  year={1997},
  publisher={American Physiological Society Bethesda, MD}
}

@article{kallet2003respiratory,
  title={Respiratory system mechanics in acute respiratory distress syndrome.},
  author={Kallet, RH and Katz, JA},
  journal={Respiratory care clinics of North America},
  volume={9},
  number={3},
  pages={297--319},
  year={2003}
}

@article{eissa1991analysis,
  title={Analysis of behavior of the respiratory system in ARDS patients: effects of flow, volume, and time},
  author={Eissa, NT and Ranieri, Vito Marco and Corbeil, C and Chass{\'e}, M and Robatto, FM and Braidy, J and Milic-Emili, J},
  journal={Journal of Applied Physiology},
  volume={70},
  number={6},
  pages={2719--2729},
  year={1991}
}

@article{farre1998respiratory,
  title={Respiratory mechanics in ventilated COPD patients: forced oscillation versus occlusion techniques},
  author={Farr{\'e}, R and Ferrer, M and Rotger, M and Torres, A and Navajas, D},
  journal={European Respiratory Journal},
  volume={12},
  number={1},
  pages={170--176},
  year={1998},
  publisher={Eur Respiratory Soc}
}

@article{guerin1993lung,
  title={Lung and chest wall mechanics in mechanically ventilated COPD patients},
  author={Guerin, C and Coussa, ML and Eissa, NT and Corbeil, C and Chasse, M and Braidy, J and Matar, N and Milic-Emili, J},
  journal={Journal of Applied Physiology},
  volume={74},
  number={4},
  pages={1570--1580},
  year={1993}
}

@article{plantier2018physiology,
  title={Physiology of the lung in idiopathic pulmonary fibrosis},
  author={Plantier, Laurent and Cazes, Aur{\'e}lie and Dinh-Xuan, Anh-Tuan and Bancal, Catherine and Marchand-Adam, Sylvain and Crestani, Bruno},
  journal={European Respiratory Review},
  volume={27},
  number={147},
  pages={170062},
  year={2018},
  publisher={Eur Respiratory Soc}
}

@article{parameswaran2006altered,
  title={Altered respiratory physiology in obesity},
  author={Parameswaran, Krishnan and Todd, David C and Soth, Mark},
  journal={Canadian respiratory journal},
  volume={13},
  number={4},
  pages={203--210},
  year={2006},
  publisher={Hindawi}
}

@article{zerah1993effects,
  title={Effects of obesity on respiratory resistance},
  author={Zerah, Fran{\c{c}}oise and Harf, Alain and Perlemuter, L{\'e}on and Lorino, Hubert and Lorino, Anne-Marie and Atlan, Guy},
  journal={Chest},
  volume={103},
  number={5},
  pages={1470--1476},
  year={1993},
  publisher={Elsevier}
}

@article{vicario2015noninvasive,
  title={Noninvasive estimation of respiratory mechanics in spontaneously breathing ventilated patients: a constrained optimization approach},
  author={Vicario, Francesco and Albanese, Antonio and Karamolegkos, Nikolaos and Wang, Dong and Seiver, Adam and Chbat, Nicolas W},
  journal={IEEE Transactions on Biomedical Engineering},
  volume={63},
  number={4},
  pages={775--787},
  year={2015},
  publisher={IEEE}
}

@article{murgu2006tracheobronchomalacia,
  title={Tracheobronchomalacia and excessive dynamic airway collapse},
  author={Murgu, Septimiu D and Colt, Henri G},
  journal={Respirology},
  volume={11},
  number={4},
  pages={388--406},
  year={2006},
  publisher={Wiley Online Library}
}

@article{colombo2011efficacy,
  title={Efficacy of ventilator waveforms observation in detecting patient--ventilator asynchrony},
  author={Colombo, Davide and Cammarota, Gianmaria and Alemani, Moreno and Carenzo, Luca and Barra, Federico Lorenzo and Vaschetto, Rosanna and Slutsky, Arthur S and Della Corte, Francesco and Navalesi, Paolo},
  journal={Critical care medicine},
  volume={39},
  number={11},
  pages={2452--2457},
  year={2011},
  publisher={LWW}
}

@article{georgopoulos2006bedside,
  title={Bedside waveforms interpretation as a tool to identify patient-ventilator asynchronies},
  author={Georgopoulos, Dimitris and Prinianakis, George and Kondili, Eumorfia},
  journal={Intensive care medicine},
  volume={32},
  number={1},
  pages={34--47},
  year={2006},
  publisher={Springer}
}

@techreport{Nagel:M382,
    Author = {Nagel, Laurence W. and Pederson, D.O.},
    Title = {SPICE (Simulation Program with Integrated Circuit Emphasis)},
    Institution = {EECS Department, University of California, Berkeley},
    Year = {1973},
    Month = {Apr},
    URL = {http://www2.eecs.berkeley.edu/Pubs/TechRpts/1973/22871.html},
    Number = {UCB/ERL M382}
}

@book{MATLAB:2019,
year = {2019},
author = {MATLAB},
title = {version 9.7.0.1216025 (R2019b) Update 1},
publisher = {The MathWorks Inc.},
address = {Natick, Massachusetts}
}

@article{yu2015simulation,
  title={Simulation of late inspiratory rise in airway pressure during pressure support ventilation},
  author={Yu, Chun-Hsiang and Su, Po-Lan and Lin, Wei-Chieh and Lin, Sheng-Hsiang and Chen, Chang-Wen},
  journal={Respiratory care},
  volume={60},
  number={2},
  pages={201--209},
  year={2015},
  publisher={Respiratory Care}
}

@article{holanda2018patient,
  title={Patient-ventilator asynchrony},
  author={Holanda, Marcelo Alcantara and Vasconcelos, Renata dos Santos and Ferreira, Juliana Carvalho and Pinheiro, Bruno Valle},
  journal={Jornal Brasileiro de Pneumologia},
  volume={44},
  number={4},
  pages={321--333},
  year={2018},
  publisher={SciELO Brasil}
}

@inproceedings{bakkes2020machine,
  title={A Machine-Learning Method for Automatic Detection and Classification of Patient-Ventilator Asynchrony},
  author={Bakkes, Tom and Montree, Roel and Mischi. Massimo and Mojoli, Francesco and Turco, Simona},
  booktitle={2020 42nd Annual International Conference of the IEEE Engineering in Medicine and Biology Society (EMBC)},
  year={2020},
  organization={IEEE}
}

@article{lino2016critical,
  title={A critical review of mechanical ventilation virtual simulators: is it time to use them?},
  author={Lino, Juliana Arcanjo and Gomes, Gabriela Carvalho and Sousa, Nancy Delma Silva Vega Canjura and Carvalho, Andrea K and Diniz, Marcelo Emanoel Bezerra and Junior, Antonio Brazil Viana and Holanda, Marcelo Alcantara},
  journal={JMIR medical education},
  volume={2},
  number={1},
  pages={e8},
  year={2016},
  publisher={JMIR Publications Inc., Toronto, Canada}
}

@article{pelosi2018close,
  title={Close down the lungs and keep them resting to minimize ventilator-induced lung injury},
  author={Pelosi, Paolo and Rocco, Patricia Rieken Macedo and de Abreu, Marcelo Gama},
  journal={Critical Care},
  volume={22},
  number={1},
  pages={72},
  year={2018},
  publisher={BioMed Central}
}

@article{marchioni2020ventilatory,
  title={Ventilatory support and mechanical properties of the fibrotic lung acting as a “squishy ball”},
  author={Marchioni, Alessandro and Tonelli, Roberto and Rossi, Giulio and Spagnolo, Paolo and Luppi, Fabrizio and Cerri, Stefania and Cocconcelli, Elisabetta and Pellegrino, Maria Rosaria and Fantini, Riccardo and Tabb{\`\i}, Luca and others},
  journal={Annals of Intensive Care},
  volume={10},
  number={1},
  pages={1--9},
  year={2020},
  publisher={SpringerOpen}
}

@article{wunsch2013icu,
  title={ICU occupancy and mechanical ventilator use in the United States},
  author={Wunsch, Hannah and Wagner, Jason and Herlim, Maximilian and Chong, David and Kramer, Andrew and Halpern, Scott D},
  journal={Critical care medicine},
  volume={41},
  number={12},
  year={2013},
  publisher={NIH Public Access}
}

@article{zhang2020detection,
  title={Detection of patient-ventilator asynchrony from mechanical ventilation waveforms using a two-layer long short-term memory neural network},
  author={Zhang, Lingwei and Mao, Kedong and Duan, Kailiang and Fang, Siqi and Lu, Yunfei and Gong, Qiang and Lu, Fei and Jiang, Ye and Jiang, Liuqing and Fang, Wenyao and others},
  journal={Computers in Biology and Medicine},
  pages={103721},
  year={2020},
  publisher={Elsevier}
}

@article{mojoli2016ventilator,
  title={Is the ventilator switching from inspiration to expiration at the right time? Look at waveforms!},
  author={Mojoli, Francesco and Iotti, Giorgio Antonio and Arnal, Jean-Michel and Braschi, Antonio},
  journal={Intensive care medicine},
  volume={42},
  number={5},
  pages={914},
  year={2016},
  publisher={Springer Science \& Business Media}
}

@article{kondili2004pattern,
  title={Pattern of lung emptying and expiratory resistance in mechanically ventilated patients with chronic obstructive pulmonary disease},
  author={Kondili, Eumorfia and Alexopoulou, Christina and Prinianakis, George and Xirouchaki, Nectaria and Georgopoulos, Dimitris},
  journal={Intensive care medicine},
  volume={30},
  number={7},
  pages={1311--1318},
  year={2004},
  publisher={Springer}
}

@article{ferreira2011sigmoidal,
  title={A sigmoidal fit for pressure-volume curves of idiopathic pulmonary fibrosis patients on mechanical ventilation: clinical implications},
  author={Ferreira, Juliana C and Bense{\~n}or, Fabio EM and Rocha, Marcelo JJ and Salge, Joao M and Harris, R Scott and Malhotra, Atul and Kairalla, Ronaldo A and Kacmarek, Robert M and Carvalho, Carlos RR},
  journal={Clinics},
  volume={66},
  number={7},
  pages={1157--1163},
  year={2011},
  publisher={SciELO Brasil}
}

@article{arbour2009cardiogenic,
  title={Cardiogenic oscillation and ventilator autotriggering in brain-dead patients: a case series},
  author={Arbour, Richard},
  journal={American Journal of Critical Care},
  volume={18},
  number={5},
  pages={496--488},
  year={2009},
  publisher={AACN}
}

@article{chao1997patient,
  title={Patient-ventilator trigger asynchrony in prolonged mechanical ventilation},
  author={Chao, David C and Scheinhorn, David J and Stearn-Hassenpflug, Meg},
  journal={Chest},
  volume={112},
  number={6},
  pages={1592--1599},
  year={1997},
  publisher={Elsevier}
}

@article{fabry1995analysis,
  title={An analysis of desynchronization between the spontaneously breathing patient and ventilator during inspiratory pressure support},
  author={Fabry, Ben and Guttmann, Josef and Eberhard, Luc and Bauer, Tilman and Haberth{\"u}r, Christoph and Wolff, Gunther},
  journal={Chest},
  volume={107},
  number={5},
  pages={1387--1394},
  year={1995},
  publisher={Elsevier}
}

@article{brown2010reference,
  title={Reference equations for respiratory system resistance and reactance in adults},
  author={Brown, Nathan J and Xuan, Wei and Salome, Cheryl M and Berend, Norbert and Hunter, Michael L and Musk, AW Bill and James, Alan L and King, Gregory G},
  journal={Respiratory physiology \& neurobiology},
  volume={172},
  number={3},
  pages={162--168},
  year={2010},
  publisher={Elsevier}
}

@article{richard2002bench,
  title={Bench testing of pressure support ventilation with three different generations of ventilators},
  author={Richard, J-C and Carlucci, A and Breton, L and Langlais, N and Jaber, S and Maggiore, S and Fougere, S and Harf, A and Brochard, L},
  journal={Intensive care medicine},
  volume={28},
  number={8},
  pages={1049--1057},
  year={2002},
  publisher={Springer}
}

@article{dres2016monitoring,
  title={Monitoring patient--ventilator asynchrony},
  author={Dres, Martin and Rittayamai, Nuttapol and Brochard, Laurent},
  journal={Current opinion in critical care},
  volume={22},
  number={3},
  pages={246--253},
  year={2016},
  publisher={Wolters Kluwer}
}
\newpage 
\appendix
\setcounter{table}{0}
\renewcommand{\thetable}{A.\arabic{table}}
\subsection{Healthy lung model equations}
\label{app:lungequations}
\begin{table}[hbt!]
\centering
 \caption{\label{tab:patienthealthyeq} Overview of the normal model equations}
 \setlength{\tabcolsep}{6pt} 
\renewcommand{\arraystretch}{1.75} 
 \scalebox{0.9}{%
 \begin{tabular}{||c c ||} 
 \hline
   & Equation \\ [0.5ex] 
    \hline\hline
chest wall volume & $V_{cw} = \frac{TLC-RV}{0.99+\exp{\frac{-(P_{cw}-A_{cw})}{B_{cw}}}}+RV$\\
Lung volume & $V_l = \frac{1}{K_l}log(\frac{P_l-B_l}{A_l})$\\
Collaspible airway volume & $V_c =\frac{V_{cmax}}{(1+e^{-A_c(P_t-B_c)})^{D_c}}$\\
Upper airway resistance & $R_u = A_u+K_u \mid \dot{V}_{cw}\mid$\\
Collapsible airway resistance & $R_c=K_c(1+e^{-A_c(P_t-B_c)})^{2D_c}$ \\
Small airway resistance & $R_s = A_s e^{K_s(V_l - RV)/(V^*-RV)}+B_s$ \\
 \hline
 \end{tabular}}
\end{table}

\subsection{Disease model equations}
\label{app:lungequationsdisease}
\setcounter{table}{0}
\renewcommand{\thetable}{B.\arabic{table}}
\begin{table}[hbt!]
\centering
 \caption{\label{tab:patientdiseaseeq} Overview of the disease model equations}
 \setlength{\tabcolsep}{6pt} 
\renewcommand{\arraystretch}{1.75} 
 \scalebox{0.9}{%
 \begin{tabular}{||c c ||} 
 \hline
   & Equation \\ [0.5ex] 
    \hline\hline
chest wall volume & $V_{cw} = \frac{TLC-RV}{0.99+\exp{\frac{-(P_{cw}-A_{cw})}{B_{cw}}}}+RV$\\
Lung volume & $V_l = \frac{A_l}{(1+e^{-B_l(P_t-
D_l)})}$\\
Collaspible airway volume & $V_c =\frac{V_{cmax}}{(1+e^{-A_c(P_t-B_c)})^{D_c}}$\\
Upper airway resistance & $R_u = A_u+K_u \mid \dot{V}_{cw}\mid$\\
Collapsible airway resistance & $R_c=K_c(1+e^{-A_c(P_t-B_c)})^{2D_c}$ \\
Small airway resistance & $R_s = A_s e^{K_s(V_l - RV)/(V^*-RV)}+B_s$ \\
 \hline
 \end{tabular}}
\end{table}

\subsection{Parameter values of the model for different patient archetypes}
\label{app:paramvalue}

\setcounter{table}{0}
\renewcommand{\thetable}{C.\arabic{table}}

\begin{table*}[hbt!]
\centering
 \caption{\label{tab:patientparam} Parameters used for the different patient archetypes}
 \scalebox{0.6}{%
 \begin{tabular}{||c c c c c c c c c c||} 
 \hline
  & Healthy & Obese 1 & Obese 2 & ARDS 1 & ARDS 2 & ARDS 3 & COPD1 & COPD2 & Idiopathic fibrosis \\ [0.5ex] 
 \hline\hline
RV (L) & 1.24 & 0.25 & 0.25 & 1.24 & 1.24 & 1.24 & 3.0 & 3.0 & 1.1 \\ 
TLC (L) & 5.19 &4.0& 4.0& 5.19& 5.19& 5.19& 7.0& 7.0& 3.4\\
Acw (cmH2O) & 1.4 & 8.0 & 8.0 & 1.4 & 1.4 & 1.4 & 3.0 & 3.0 & 1.4\\
Bcw (cmH2O) & -3.5 & -5.0 & -5.0 & -3.5 & -3.5 & -3.5 & -5.0 & -5.0 & -3.0\\
Al & 0.2 & 3.0 & 3.0 & 3.7 & 3.0 & 2.3 & 6.9 & 6.9 & 3.4\\
Bl & -0.5 & 0.2 & 0.15 & 0.15 & 0.15 & 0.15 & 0.5 & 0.5 & 0.06\\
Dl  & 0 &10.0& 14.4& 10.0& 14.4& 20.0& 2.3& 2.3& 7.0\\
Kl & 1.0 &0& 0& 0& 0& 0& 0& 0& 0\\ 
As (cmH2O/(L.s))& 2.2 &4.0& 7.0& 3.0& 7.0& 18.0& 2.2& 7.0& 2.0\\
Bs (cmH2O/(L.s))& 0.02& 0.5& 1.5& 1.0& 1.5& 2.0& 1.5& 3.0& 0.5\\ 
Ks & -10.9& -5.0& -5.0& -4.0& -5.0& -14.0& -20.0& -2.5& -10.9\\
Vstar (L) & 5.3 &4.0& 3.5& 4.5& 3.5& 3.0& 5.3& 6.0& 3.4\\
Vcmax (L) & 0.1 &0.07 &0.06 &0.07 &0.06 &0.05 &0.05 &0.05 &0.1\\
Ac (cmH2O) & 0.341 & 0.4 & 0.55 & 0.6 & 0.55 & 0.77 & 1.5 & 1.2 & 0.341\\
Bc (cmH2O) & 9.692 &9.0& 10.0& 10.0& 17.0& 27.0& 6.2& 7.0& 7.0\\
Dc & 0.411& 0.411& 0.411& 0.411& 0.411& 0.411& 0.411& 0.411& 0.411\\
Kc &0.21& 0.5& 1.5& 1.0& 1.5& 2.4& 2.0& 0.6& 0.5\\
Au (cmH2O/(L.s)) & 0.34 &2.0& 2.0& 0.34& 0.34& 0.34& 0.34& 0.34& 0.34\\
Ku (cmH2O/(L\textsuperscript{2}.s\textsuperscript{2})) & 0.46 &8.0& 8.0& 0.46& 0.46& 0.46& 0.46& 2.5& 0.46\\
Rd ((cmH2O.s)/L))& 1.0 & 18.0 & 18.0 & 3.0 & 3.0 & 3.0 & 3.0 & 1.0 & 2.0\\
Cd (L/cmH2O) & 0.5& 0.2& 0.2& 0.2& 0.2& 0.2& 0.2& 0.5& 0.05\\ [1ex] 
 \hline
 \end{tabular}}
\end{table*}
\end{document}